%% file: neurips_2026.tex
\documentclass{article}

\usepackage[preprint]{neurips_2026}
\usepackage{graphicx}
\usepackage{amsmath}
\usepackage{float}
\usepackage{wrapfig}
\usepackage{caption}
\usepackage{booktabs}
\usepackage{graphicx}

\usepackage[utf8]{inputenc} % allow utf-8 input
\usepackage[T1]{fontenc}    % use 8-bit T1 fonts
\usepackage{hyperref}       % hyperlinks
\usepackage{url}            % simple URL typesetting
\usepackage{booktabs}       % professional-quality tables
\usepackage{amsfonts}       % blackboard math symbols
\usepackage{nicefrac}       % compact symbols for 1/2, etc.
\usepackage{microtype}      % microtypography
\usepackage{xcolor} 
\usepackage[most]{tcolorbox}
\usepackage{xcolor}
\usepackage{booktabs}
\usepackage{adjustbox}
\usepackage{amsmath}% colors
\definecolor{prefixblue}{RGB}{235,245,255}
\definecolor{abranchred}{RGB}{255,241,241}
\definecolor{bbranchgreen}{RGB}{239,250,242}
\definecolor{contgray}{RGB}{248,248,248}

\newtcolorbox{modelbox}[2][]{
  enhanced,
  breakable,
  colback=#2,
  colframe=black!35,
  boxrule=0.35pt,
  arc=1.5mm,
  left=1mm,
  right=1mm,
  top=1mm,
  bottom=1mm,
  fontupper=\footnotesize,
  #1
}

\title{It Takes 8 Tokens: Weak-to-Strong Off-Policy RL via Auxiliary Branches}

\author{
    \textbf{Dayu Wang}$^{1,2}$ \quad
    \textbf{Jiaye Yang}$^{1}$ \quad
    \textbf{Weikang Li}$^{1}$\thanks{Project leads and corresponding authors.} \quad
    \textbf{Jiahui Liang}$^{1}$ \quad
    \textbf{Liwei Qian}$^{1}$ \quad
    \textbf{Xin Pei}$^{1}$ \quad
    \textbf{Jizhou Huang}$^{1}$\footnotemark[1] \\
    $^{1}$Baidu Inc. \qquad
    $^{2}$Nanyang Technological University \\
    \texttt{dayu001@e.ntu.edu.sg} \quad
    \texttt{yamseyoung@gmail.com} \\
    \texttt{wavejkd@pku.edu.cn} \quad
    \texttt{\{liangjiahui03, qianliwei, Xin Pei,huangjizhou01\}@baidu.com}
}

\newcommand{\method}{W2SPO}
\begin{document}

\maketitle

\begin{abstract}
Reinforcement learning with verifiable rewards has emerged as a standard approach for enhancing reasoning in large language models, which typically optimizes the policy by contrasting multiple self-generated rollouts. However, we identify a critical \textbf{support-limited} bottleneck in this paradigm: on challenging reasoning tasks, the target model's samples often exhibit semantic redundancy, converging into the same erroneous "reasoning basins" that offer negligible reward contrast for policy updates.
In this paper, we propose to overcome this limitation through a \textbf{weak-to-strong learning} paradigm, where a policy's exploration is informed by a weaker but computationally efficient auxiliary model.  
We introduce \method{}, an off-policy RL method that injects short auxiliary segments—often as brief as 8 tokens—into intermediate target-model trajectories and the target model then completes the reasoning path from these diverted states. Policy updates are restricted to these short inserted segments based on final verifiable rewards. 
Empirically, \method{} achieves superior performance among evaluated 4B-scale models on mathematical reasoning benchmarks, outperforming evaluated post-trained baselines. Compared with vanilla GRPO under the same sampling budget, \method{} improves Pass@1 from 62.3\% to 64.2\% while achieving a \textbf{3.55$\times$} training speedup. These results suggest that weak auxiliary branches can induce stronger target reasoning policies by expanding local exploration support. Code and data are available at \url{https://anonymous.4open.science/r/W2SPO-1B86/}.
\end{abstract}

\section{Introduction}
\begin{figure}[t]
    \centering
    \includegraphics[width=1\linewidth]{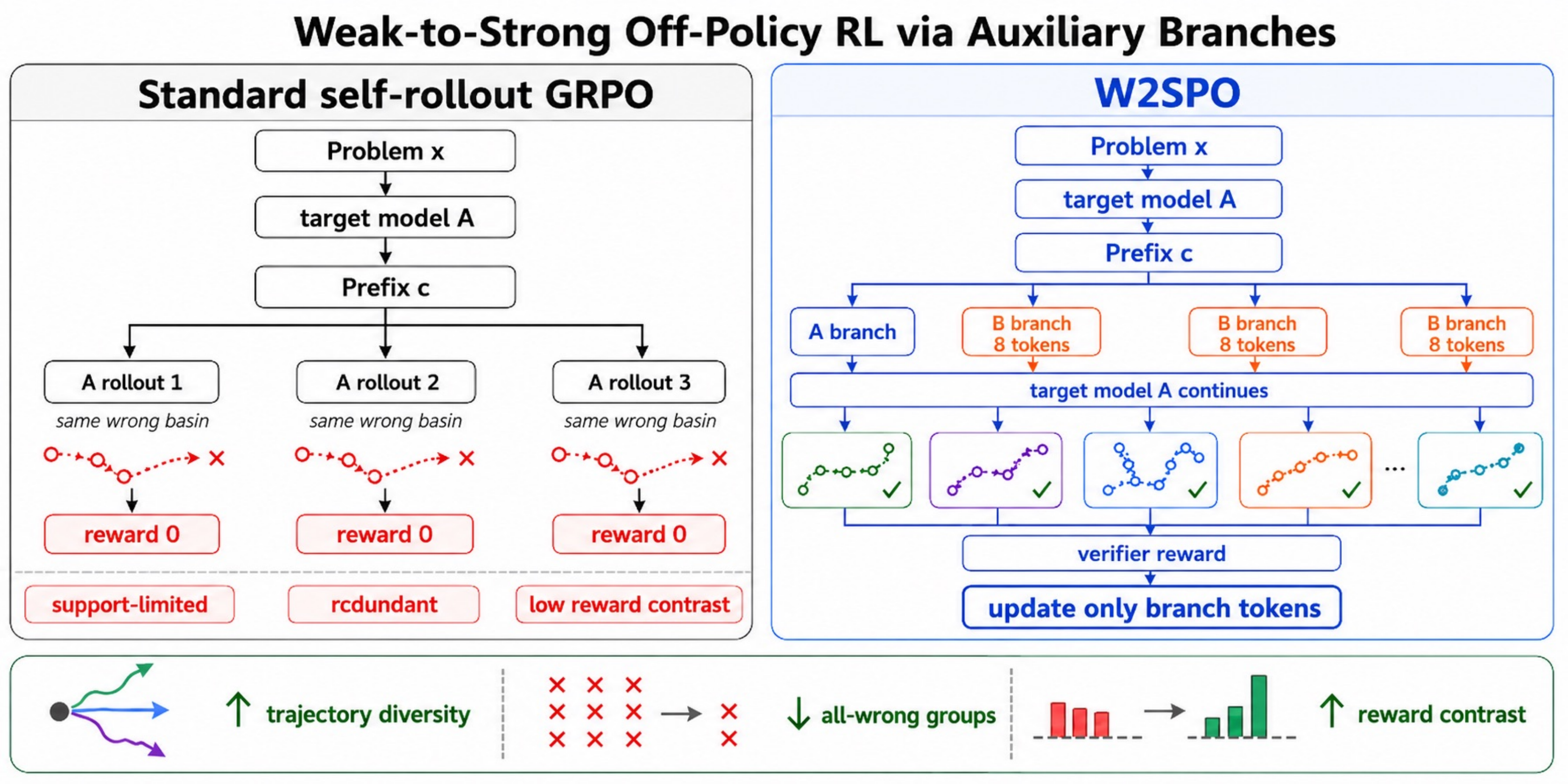}
    \caption{Overview of \method{}. 
Standard GRPO relies on self-rollouts from the target model, which can be semantically redundant and yield all-wrong rollout groups on difficult reasoning problems. 
\method{} locally expands the rollout support by inserting short auxiliary branches into an intermediate target-model prefix, then lets the target model complete the solution. 
Verifier rewards are assigned to the completed trajectories, while policy updates are applied only to the short branch segment.
}
    \label{fig:method}
\end{figure}
Reinforcement learning with verifiable rewards has become a central approach for improving language-model reasoning~\citep{shao2024deepseekmathpushinglimitsmathematical,Guo_2025,song2025outcomebasedexplorationllmreasoning}. Methods such as PPO- and GRPO-style training optimize models from sampled reasoning traces using outcome rewards, allowing models to discover alternative solution paths without imitating a fixed supervised trace~\citep{wen2025reinforcementlearningverifiablerewards,cobbe2021trainingverifierssolvemath}. However, these methods rely heavily on self-generated rollouts from the target model. In standard GRPO, the current policy samples multiple trajectories for the same problem and updates the model according to relative advantages within the group. This procedure assumes that the sampled rollouts are sufficiently diverse and informative~\citep{wang2023selfconsistencyimproveschainthought}. For difficult reasoning problems, this assumption often fails: target-model rollouts may look lexically different but follow similar intermediate logic, enter the same erroneous reasoning basin, and fail in similar ways~\citep{dang2025assessing}. Thus, the bottleneck is not merely insufficient sampling budget, but limited coverage of the rollout distribution~\citep{yue2025does,brown2024large}.

This support limitation weakens reasoning RL in two ways~\citep{dong2026rlpluscounteringcapabilityboundary}. When all rollouts are incorrect, the group provides little useful reward contrast~\citep{chen2026stepwiseguidedpolicyoptimization}. Even when rewards differ, the variation may remain within a narrow region of the target model's current reasoning distribution, causing RL to reinforce existing behaviors rather than discover low-probability reasoning turns that could redirect the trajectory toward a correct solution~\citep{li2026back}. This motivates a different form of exploration: instead of simply increasing the number of self-rollouts, can we locally expand the target model's rollout support while still letting the target model complete the long-horizon reasoning process?

We study a weak-to-strong alternative~\citep{burns2023weaktostronggeneralizationelicitingstrong,wang2026student}. Instead of using an auxiliary model as a teacher, we use it as a local exploration proposer~\citep{hsieh2023distilling}. Our key observation is that a weaker auxiliary model can still provide useful short reasoning segments, even when its full solutions are inferior to those of the target model~\citep{lee2024can}. Such segments can perturb the target model away from its dominant reasoning pattern and into a different continuation basin~\citep{chia2024reasoning}. After the segment is inserted, the target model resumes generation and remains responsible for completing the long-horizon reasoning path. The auxiliary model therefore does not need to solve the problem; it only needs to propose local branches that expand the target model's exploration support~\citep{yao2023tree}.

Motivated by this observation, we propose \method{}, a Weak-to-Strong Off-Policy RL method via Auxiliary Branches for reasoning models~\citep{hou2025treerl}. Given a problem, the target model first generates an intermediate reasoning prefix. We then sample short branches from both the target model and an auxiliary model at this intermediate state~\citep{wei2022chain}. In our main instantiation, each auxiliary branch contains only 8 tokens. The target model continues from the inserted branch to produce a complete solution, which is scored by a verifier or rule-based reward~\citep{wang2024math}. The resulting reward estimates the value of the local branch, and the policy update is applied to the branch segment rather than the entire continuation~\citep{setlur2024rewarding}.

Our contributions are summarized as follows:
\begin{itemize}
    \item \textbf{Observation.} We identify a support-limitation failure mode in self-rollout reasoning RL: target-model rollouts can be semantically redundant, remain trapped in the same erroneous reasoning basins, and provide insufficient reward contrast for GRPO-style optimization.

    \item \textbf{Method.} We propose \method{}, an auxiliary-branched RL framework that uses a weaker model as a local proposal distribution. The method inserts short auxiliary branches into intermediate target-model trajectories, lets the target model complete the remaining reasoning path, and performs importance-aware policy updates on the local branch segment.

    \item \textbf{Empirical findings.} We show that short auxiliary branches, instantiated as 8-token segments, improve reasoning RL over vanilla GRPO and target-only branching. They increase distinct reasoning trajectories and correct-solution diversity, reduce all-wrong rollout groups.
\end{itemize}

\section{Related Work}

\paragraph{Reasoning RL with verifiable rewards.}
Recent reasoning models increasingly rely on reinforcement learning with verifiable rewards, where correctness can be judged by rules, symbolic checkers, or learned verifiers~\citep{shao2024deepseekmathpushinglimitsmathematical,Guo_2025,song2025outcomebasedexplorationllmreasoning,wen2025reinforcementlearningverifiablerewards,cobbe2021trainingverifierssolvemath}. GRPO-style methods compare multiple sampled responses to the same problem and update the policy using relative rewards, making them effective for mathematical and programmatic reasoning. However, their effectiveness depends on whether the sampled rollout group contains diverse and useful solution attempts. Prior work on self-consistency shows that multiple trajectories can improve answer reliability when they cover meaningfully different reasoning modes~\citep{wang2023selfconsistencyimproveschainthought}. On difficult problems, however, repeated samples may remain concentrated in the same reasoning basin, providing limited reward contrast despite surface-level variation~\citep{dang2025assessing,yue2025does,brown2024large}. Our work focuses on this support-limited regime and studies how to make rollout groups more informative for policy optimization.

\paragraph{External guidance, distillation, and branching.}
A common way to improve exploration is to introduce external guidance~\citep{deng2025supervised}. Supervised distillation and expert imitation can provide high-quality reasoning traces, but they require strong teachers or curated solutions and may suffer from mismatch between expert traces and the target model's native reasoning style~\citep{kim2025their}. Expert-guided branching methods insert partial hints into reasoning trajectories to help models continue from more promising intermediate states~\citep{zhang2025bread,wang2025hint}. These methods demonstrate the value of intermediate guidance, but they often rely on expert or ground-truth traces. In contrast, \method{} does not assume access to complete expert solutions. It uses a weaker auxiliary model only as a local proposal distribution, while the target model remains responsible for completing the final reasoning trajectory.

\paragraph{Weak-to-strong learning.}
Our method is closely related to weak-to-strong learning and model collaboration~\citep{burns2023weaktostronggeneralizationelicitingstrong,wang2026student,hsieh2023distilling,lee2024can}. Unlike traditional distillation where a strong teacher supervises a weak student, recent evidence suggests that even a weaker model can effectively guide a stronger one by providing diverse local interventions~\citep{wang2026student}. This is in line with our observation of local complementarity: a short segment from a weaker auxiliary model can successfully nudge the target model into a different reasoning basin that it might otherwise fail to explore~\citep{chia2024reasoning}. By treating the auxiliary model as a source of local perturbations rather than a global solver, \method{} expands the support of the rollout distribution while preserving the target model's superior long-form reasoning capabilities~\citep{nath2025adaptive}. By restricting updates to reasoning over these short inserted segments, we localize credit assignment to the critical decision points that shift downstream trajectories~\citep{roux2025tapered,guo2025segment}. Thus, \method{} uses heterogeneous short branches to construct more informative RL rollout groups rather than imitating full auxiliary trajectories~\citep{yan2025learning,lyu2025correction,liu2025uniform}.

\section{Motivation}

Modern reasoning RL methods with verifiable rewards, such as GRPO, typically rely on multiple self-generated rollouts by the current policy and optimize the model using group-relative reward differences~\citep{shao2024deepseekmathpushinglimitsmathematical}. 
This paradigm implicitly assumes that increasing the number of sampled trajectories will produce not only more attempts, but also more informative reasoning alternatives. 
Such an assumption is consistent with the intuition behind self-consistency, where sampling diverse reasoning paths can improve answer reliability~\citep{wang2022self}. 
However, on difficult reasoning problems~\citep{aime24,aime25,he2024olympiadbench,hendrycks2021measuring}, we observe a clear diminishing-return effect from increased self-sampling: as the sampling budget grows, the proportion of correct trajectories that are also semantically distinct becomes progressively smaller.

\begin{figure}[H]
    \centering
    \includegraphics[width=1\linewidth]{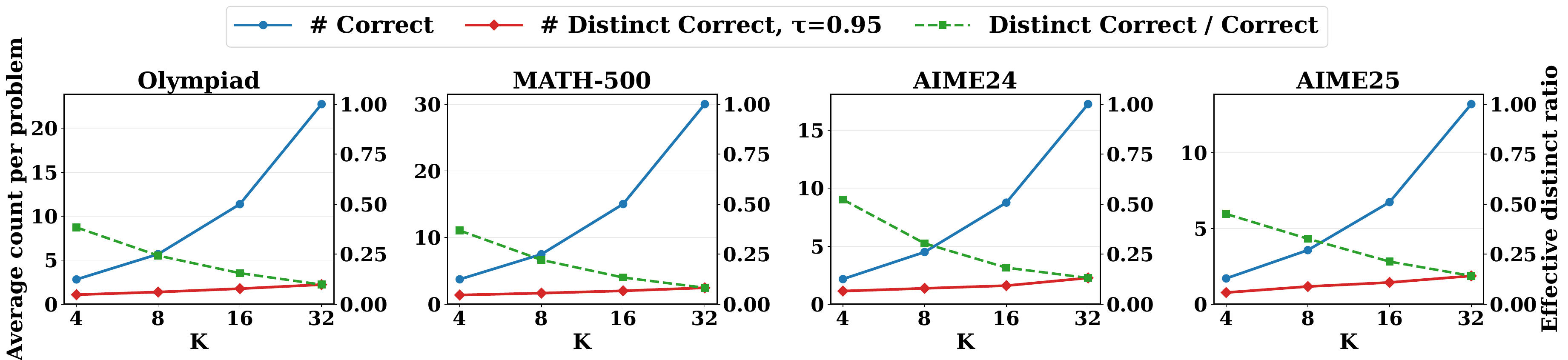}
    \caption{
    Self-rollout is diminishing-return. 
    As the sampling budget $K$ increases (temperature=0.7, top-p=0.95), the total number of correct samples increases, but the number of semantically distinct correct solutions grows much more slowly. Sensitivity analysis is at~\ref{sec:sensitivity_analysis}. 
    }
    \label{fig:self_rollout_support}
\end{figure}

\paragraph{Failure mode 1: increasing self-sampling does not necessarily increase effective diversity.}
For each problem, we sample $K \in \{8,16,32\}$ trajectories from the target model $A$ (Qwen3-4B-Instruct-2507~\citep{qwen3technicalreport}) and measure both final-answer correctness and semantic diversity. 
Semantic diversity is computed by embedding (bge-small-en-v1.5~\citep{bge_embedding}) generated reasoning traces and counting semantically distinct trajectories under a cosine-similarity threshold $\tau=0.95$. 
As shown in Figure~\ref{fig:self_rollout_support}, increasing $K$ substantially increases the number of sampled trajectories and often the total number of correct samples. 
However, the number of semantically distinct correct solutions grows much more slowly, and the effective distinct-correct ratio decreases as $K$ increases. 
This suggests a clear diminishing-return effect in self-sampling: simply increasing the self-sampling budget is often not an efficient way to expand the effective reasoning support of the target model, as additional samples mainly introduce surface-level variations rather than genuinely new reasoning modes.

This observation is aligned with recent findings on reasoning diversity and RLVR. 
Repeated sampling can improve coverage when the model's sampling distribution already contains correct solutions~\citep{brown2024large}; however, recent studies show that reasoning models may suffer from diversity collapse, where fine-tuning or RL concentrates probability mass on a narrower set of reasoning paths~\citep{dang2025assessing}. 
Outcome-based RL has also been shown to induce systematic losses in generation diversity, which can undermine test-time scaling~\citep{song2025outcome}. 
Moreover, analyses of RLVR suggest that current RL methods often improve sampling efficiency by reweighting existing reasoning paths rather than eliciting fundamentally new reasoning patterns beyond the base model's support~\citep{yue2025does}. 
These results indicate that the key bottleneck is not merely the number of rollouts, but whether the rollout group covers sufficiently different reasoning modes.

\paragraph{Failure mode 2: hard problems produce all-wrong groups and no effective update.}
The limited support of self-rollouts is less effective for RL training. 
For problems that the target model cannot solve under its current rollout distribution, all sampled responses may receive zero verifier reward. 
In GRPO-style objectives, such all-wrong groups have little or no reward contrast, leading to zero or near-zero group-relative advantages and therefore no effective policy-gradient update for those prompts. 
Recent work on zero-advantage or negative rollout groups similarly observes that when all responses in a group receive identical rewards, standard GRPO wastes sampling compute and provides no useful gradient signal~\citep{feng2025don}. 
Thus, increasing the self-rollout budget may still fail to train on the hardest examples if the model cannot sample any successful trajectory within the group.

\paragraph{Auxiliary branches as local support expansion.}
To address these two limitations, we do not replace the target model with a stronger teacher or imitate a full auxiliary solution. 
Instead, we use an auxiliary model $B$ (gemma-3-4B-IT~\citep{gemma_2025}) as a local proposal distribution: given an intermediate prefix generated by the target model $A$, $B$ proposes a short branch segment, and then $A$ resumes generation to complete the solution. 
This forms an \textsc{A-B-A} trajectory: target prefix, auxiliary branch, and target continuation. 
Thus, the role of $B$ is not to solve the problem end-to-end, but to inject local reasoning turns that may redirect $A$ away from its dominant erroneous basin and into a more productive continuation region.

\begin{figure}[t]
    \centering

    \begin{minipage}[c]{0.60\linewidth}
        \centering
        \small
        \setlength{\tabcolsep}{5pt}
        \renewcommand{\arraystretch}{1.12}
        \begin{tabular}{lccccc}
            \toprule
            \textbf{Method} &
            \textbf{Olympiad} &
            \textbf{MATH500} &
            \textbf{AIME24} &
            \textbf{AIME25}  \\
            \midrule
            Direct
            & 2.23 & 2.46 & 2.27 & 1.87 &\\
            A-B-A
            & \textbf{2.49} & \textbf{2.73} & \textbf{2.30} & \textbf{2.17} \\
            \midrule
            Rel. Gain
            & +12.0\% & +11.1\% & +1.5\% & +16.1\%  \\
            \bottomrule
        \end{tabular}
    \end{minipage}
    \hfill
    \begin{minipage}[c]{0.36\linewidth}
        \centering
        \includegraphics[width=\linewidth]{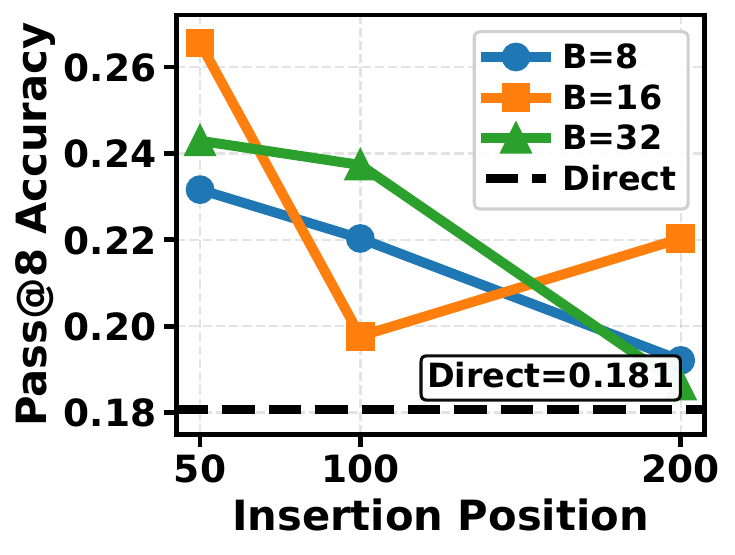}
    \end{minipage}

    \caption{
    Auxiliary branches expand effective reasoning support.
    Left: average number of semantically distinct correct trajectories under direct self-rollout and \textsc{A-B-A} inference. 
    Right: Pass@8 performance under different insertion positions and lengths. All samples are generated with temperature=0.7, top-p=0.95, and the same system prompt.
    }
    \label{fig:aba_diversity_and_position}
\end{figure}

Under the same sampling budget, this auxiliary-branch construction yields more diverse and effective reasoning trajectories. 
As shown in Figure~\ref{fig:aba_diversity_and_position}, \textsc{A-B-A} produces a higher average number of distinct correct trajectories than direct self-rollout inference, supporting the view that short auxiliary branches expand the local continuation support of the target model rather than merely increasing the number of samples. 
To further test this mechanism on hard problems, we construct a subset where direct sampling from $A$ fails under Pass@4, and evaluate different intervention positions and branch lengths. 
Auxiliary branches consistently outperform the direct Pass@8 baseline, with the strongest gains from early interventions: the direct baseline achieves only $18.08\%$ Pass@8, while inserting a branch at position 50 improves performance to $23.16\%$, $26.55\%$, and $24.29\%$ with branch lengths of 8, 16, and 32 tokens, respectively. 
Since the 8-token branch at position 50 already provides substantial improvement with the least off-policy context and auxiliary-generation cost, we use this configuration in subsequent reinforcement learning experiments.

\paragraph{Design implication.}
These findings motivate a branch-level view of reasoning exploration: a short externally proposed segment inserted near the early reasoning stage can already reshape the target model's continuation distribution, increasing subsequent-sample diversity and improving success on hard problems where target-only self-rollouts often fail.

We therefore train the target model to internalize this ability while retaining its long-horizon reasoning behavior. 
Our method combines off-policy auxiliary-branch sampling with reinforcement learning over a short injected window: the auxiliary model supplies short branch proposals, the target model completes the trajectory, and the final verifier reward assigns credit only to the inserted branch tokens. 
This expands local exploration support without distilling full auxiliary trajectories, keeps the update focused on the decision point that changes downstream reasoning, and updates only the target model $A$ while keeping the auxiliary model $B$ fixed as an external proposal distribution. 
The next section formalizes this branch-level off-policy RL objective.

\section{Method}

We propose a Weak-to-Strong Off-Policy RL method for improving LLM reasoning, as shown in Figure~\ref{fig:method}. 
The method consists of five steps: greedy prefix generation, short branch proposal, rollout-based branch scoring, importance-weighted advantage estimation, and clipped policy optimization.

\paragraph{Step 1: Greedy prefix generation.}
For each prompt $x$, the actor first generates a reasoning prefix $z$ of length $L_p$ under greedy decoding:
\[
z \sim \pi_{\theta_{\mathrm{old}}}(\cdot \mid x).
\]
This prefix represents the early reasoning context of the target model and serves as the intervention point for subsequent branch exploration.

\paragraph{Step 2: Short branch proposal.}
Starting from the prefix $(x,z)$, we construct short candidate branches from two sources. 
The actor samples $n_A$ branches,
\[
b_i^A \sim \pi_{\theta_{\mathrm{old}}}(\cdot \mid x,z),
\qquad i=1,\ldots,n_A,
\]
while a frozen auxiliary model samples $n_B$ textual continuations,
\[
\tilde{b}_j^B \sim q_B(\cdot \mid x,z),
\qquad j=1,\ldots,n_B.
\]
Since the actor and auxiliary model may use different tokenizers, auxiliary continuations are decoded into text and re-encoded by the actor tokenizer. 
All branches are truncated or padded to the same actor-token length $L_b$. 
In our configuration, $L_p=50$, $L_b=8$, $n_A=2$, and $n_B=6$, giving eight candidate branches per prompt.

\paragraph{Step 3: Rollout-based branch scoring.}
Each branch $b_i$ is evaluated by letting the actor continue from the branched context $(x,z,b_i)$ for $K$ rollouts:
\[
u_{i,k} \sim \pi_{\theta_{\mathrm{old}}}(\cdot \mid x,z,b_i),
\qquad k=1,\ldots,K.
\]
A task-specific rule-based reward function compares the completed solution with the ground-truth label $y^\star$. 
The branch reward is defined as the mean rollout reward:
\[
r_i = \frac{1}{K}\sum_{k=1}^{K}
R(x,z,b_i,u_{i,k},y^\star).
\]
Therefore, a branch is not judged by its local text alone, but by whether it leads the actor toward successful future reasoning. 
During training, the action mask is applied only to the branch tokens.

\paragraph{Step 4: Importance-weighted branch advantage.}
Because branches are sampled from a mixture of the actor and the auxiliary model, we compute an importance-weighted group advantage. 
Let
\[
p_i =
\sum_{t=1}^{L_b}
\log \pi_{\theta_{\mathrm{old}}}(b_{i,t}\mid x,z,b_{i,<t})
\]
be the old actor log-probability of branch $b_i$, and let $q_i$ denote the proposal log-probability from its sampling source. 
For auxiliary branches, $q_i$ is the auxiliary log-probability length-normalized after re-tokenization.

The importance coefficient is
\[
c_i
=
\min\left\{
c_{\max},
\exp\left(
\alpha\cdot
\operatorname{clip}
\left(
p_i-q_i,
-30,30
\right)
\right)
\right\},
\]
where we use $\alpha=0.02$ and $c_{\max}=2.0$.

For each prompt group, we compute the weighted mean and variance:
\[
\mu = \frac{\sum_i c_i r_i}{\sum_i c_i},
\qquad
\sigma =
\sqrt{
\frac{\sum_i c_i(r_i-\mu)^2}{\sum_i c_i}+\epsilon
}.
\]
A leave-one-out weighted baseline is then used:
\[
b_i =
\frac{\sum_j c_j r_j - c_i r_i}
{\sum_j c_j - c_i + \epsilon}.
\]
The final branch-level advantage is
\[
A_i = \frac{c_i(r_i-b_i)}{\sigma},
\]
which is clipped to $[-3,3]$. 
Prompt groups with zero reward variance are assigned zero advantage.

\paragraph{Step 5: Clipped policy update on branch tokens.}
Finally, the actor is optimized using the PPO clipped objective, but gradients are applied only to the masked branch tokens. 
For each branch token, we define
\[
\rho_{i,t}(\theta)
=
\exp\left(
\log \pi_\theta(b_{i,t}\mid x,z,b_{i,<t})
-
\log \pi_{\theta_{\mathrm{old}}}(b_{i,t}\mid x,z,b_{i,<t})
\right).
\]
The policy loss is
\[
\mathcal{L}_{\mathrm{policy}}
=
-\mathbb{E}_{i,t}
\left[
\min
\left(
\rho_{i,t}(\theta)A_i,
\operatorname{clip}(\rho_{i,t}(\theta),1-\epsilon,1+\epsilon)A_i
\right)
\right],
\]
with $\epsilon=0.2$. 
Since the PPO clipping constraint already provides a relatively strict bound on policy updates, we do not use an additional KL-divergence regularization term; supervision comes directly from the rule-based rollout reward. 
Thus, the update explicitly increases the actor's probability of short reasoning branches that lead to correct final solutions, while leaving the prefix and continuation tokens unoptimized. An example rollout is provided in Appendix ~\ref{subsec:example_rollout}.

\begin{table*}[h]
\centering
\caption{
Main greedy decoding performance and runtime comparison. 
Top: greedy decoding performance measured by Pass@1. 
Bottom: runtime comparison between matched-sampling GRPO and W2SPO(ours) under the same sampling budget and sampling hyperparameters.
Runtime measurements are collected on 8$\times$A800 GPUs, and we report per-step mean$\pm$std in seconds.
}
\label{tab:main_results_and_runtime}

\small
\textbf{(a) Greedy decoding performance measured by Pass@1.}

\vspace{0.4em}
\resizebox{\textwidth}{!}{
\begin{tabular}{lcccccc}
\toprule
\textbf{Model} &
\textbf{Overall} &
\textbf{MATH500} &
\textbf{Olympiad} &
\textbf{Omni-HARD} &
\textbf{AIME24} &
\textbf{AIME25} \\
\midrule
Qwen3-4B-Instruct-2507 &
58.3 & 90.8 & 64.4 & 21.2 & 50.0 & 40.0 \\
Gemma-3-4B-IT &
40.7 & 75.2 & 39.9 & 13.4 & 3.3 & 10.0 \\
ThinkTwice-Qwen3-4B-Instruct &
59.4 & 92.2 & 66.6 & 22.1 & 36.7 & 30.0 \\
Qwen3-4B-DAPO-math-reasoning &
59.8 & 93.4 & 65.7 & 23.1 & 36.7 & 40.0 \\
Frugal-Thinking-4B &
61.4 & 92.8 & 69.3 & 23.3 & 56.7 & 40.0 \\
NPR-4B &
52.0 & 86.0 & 57.6 & 15.1 & 40.0 & 23.3 \\
Polaris-4B-Preview &
43.4 & 78.2 & 44.8 & 10.6 & 30.0 & 20.0 \\
Qwen3-4B-Reinforce-Ada &
61.1 & 92.0 & 69.1 & 23.6 & 43.3 & 43.3 \\
Matched-sampling GRPO &
62.3 & 92.6 & 69.3 & 26.5 & 53.3 & 40.0 \\
\textbf{W2SPO (ours)} &
\textbf{64.2} & 93.2 & \textbf{71.8} & \textbf{28.4} &
\textbf{56.7} & \textbf{46.7} \\
\bottomrule
\end{tabular}
}

\vspace{1.0em}

\textbf{(b) Runtime breakdown under matched sampling budget.}

\vspace{0.4em}
\begin{tabular}{lcccc}
\toprule
\textbf{Method} &
\textbf{Process (s)} &
\textbf{Forward (s)} &
\textbf{Train (s)} &
\textbf{Total (s)} \\
\midrule
GRPO &
$782.22 \pm 67.08$ &
$81.44 \pm 9.41$ &
$945.11 \pm 34.04$ &
$1808.78 \pm 89.98$ \\
W2SPO(ours) &
$482.61 \pm 35.81$ &
$2.09 \pm 0.29$ &
$24.40 \pm 0.78$ &
$509.11 \pm 35.91$ \\
\midrule
\textbf{Speedup} &
$\mathbf{1.62\times}$ &
$\mathbf{38.97\times}$ &
$\mathbf{38.73\times}$ &
$\mathbf{3.55\times}$ \\
\bottomrule
\end{tabular}

\end{table*}

\section{Experiments}

\subsection{Training Setup}
In our experiment, the trainable actor model $A$ is initialized from Qwen3-4B-Instruct-2507, while an
auxiliary model $B$ initialized from Gemma-3-4B-IT is used only as a frozen off-policy branch proposer. 
%Results of various model pairs are presented in Table \ref{tab:robustness_analysis}. 
Our training data is sampled from Skywork-OR1-RL-Data~\citep{he2025skywork,skywork-or1-2025}, with all problems identical or overlapping with the evaluation benchmarks removed from the training set. More details on training experiments can be found in Appendix~\ref{app:exp_details}. %We evaluate our method on mathematical reasoning tasks. 

% The training prompts are
% sampled from a JSONL dataset, where each example contains a problem prompt and a
% ground-truth answer label. The prompt is formatted with the tokenizer chat
% template before being fed into the model. Unless otherwise specified, the actor
% model is initialized from Qwen3-4B-Instruct-2507, and the auxiliary branch
% proposer is initialized from Gemma-3-4B-IT. The actor is trained with the
% auxiliary-branch GRPO objective described in the previous section. The auxiliary
% model is kept frozen throughout training and is only used to propose short
% candidate branches.

% Training is conducted with Ray-based distributed rollout generation and
% DeepSpeed-based actor optimization. We use six vLLM engines for generation and
% two GPUs for actor training. The maximum prompt length is set to $2024$, and the
% maximum generation length is set to $8192$. For each prompt, we sample
% $n_A=2$ actor branches and $n_B=6$ auxiliary branches, resulting in eight
% branches per prompt. The prefix length is $50$ tokens, the branch length is
% $8$ tokens, and each branch is evaluated with two continuation rollouts. The
% importance-weighted advantage estimator uses $\alpha=0.02$, $c_{\max}=2.0$,
% and advantage clipping threshold $3.0$. The PPO clipping coefficient is set to
% $0.2$. We train for two episodes with a learning rate of $10^{-6}$ and cosine
% learning-rate scheduling. Since the auxiliary-branch setting directly uses a
% rule-based reward function, the reference model, reward model, and KL penalty
% are disabled during this training stage.

\subsection{Evaluation Protocol}
We evaluate models on a collection of mathematical reasoning benchmarks,
including MATH500, Olympiad-test, Omni-HARD~\citep{gao2024omnimathuniversalolympiadlevel,Polaris2025}, AIME 2024, and AIME
2025. The evaluation set contains $1763$ problems in total. For each generated solution, correctness is determined using a combination of regular-expression-based answer extraction and the \texttt{math-verify} toolkit. All evaluations are conducted under the same system prompt, conversation format, and context-length setting to ensure a fair comparison across methods.

We compare our method with both pretrained open-source models and training-based
approaches. We include Qwen3-4B-Instruct-2507 as the direct initialization baseline. This
baseline measures the performance before reinforcement learning. We also compare \method{} against recent open-source 4B reasoning models, including 
models trained with supervised reasoning distillation, RLVR, and adaptive RL sampling~\citep{jiao2026thinktwice,yu2025dapo,bounhar2025shorter,wu2025native,xiong2025reinforce}. 
All baselines are evaluated under the same decoding and verification protocol.

To isolate the effect of auxiliary branch construction from the total number of
samples, we include a standard GRPO baseline with the same number of samples per
prompt. This baseline uses the same actor initialization and the same overall
rollout budget.

\subsection{Main Results}

Table~\ref{tab:main_results_and_runtime} details the greedy decoding performance and runtime efficiency. Our method is expected to improve the actor's most likely reasoning path by assigning higher probability to short branches that lead to successful completions. Compared with the base model, this evaluates whether auxiliary-branch training improves the deterministic solution quality. Compared with matched-sampling GRPO and the A-only branch ablation, it evaluates whether the improvement comes from the proposed auxiliary branch mechanism rather than from additional sampling alone.

As shown in Table~\ref{tab:main_results_and_runtime} (a), our method achieves strong Pass@1 accuracy across all tested models, including specialized reasoning models like Qwen3-4B-Reinforce-Ada. Crucially, when compared against the matched-sampling GRPO, our approach yields superior Pass@1 scores across all benchmarks (Overall score 64.2 vs. 62.3), particularly in challenging sets like Omni-HARD (+1.9\%) and AIME25 (+6.7\%). This confirms that the performance gain stems from the structural merit of the auxiliary branch mechanism rather than a mere increase in sampling volume. Furthermore, the runtime analysis in Table~\ref{tab:main_results_and_runtime} (b) highlights that our method is not only more accurate but also substantially faster, achieving a 38.97$\times$ acceleration in forward passes and a 3.55$\times$ total speedup.

\begin{wrapfigure}[12]{r}{0.4\linewidth}
    \centering
    \vspace{-2.5em}
    \includegraphics[width=\linewidth]{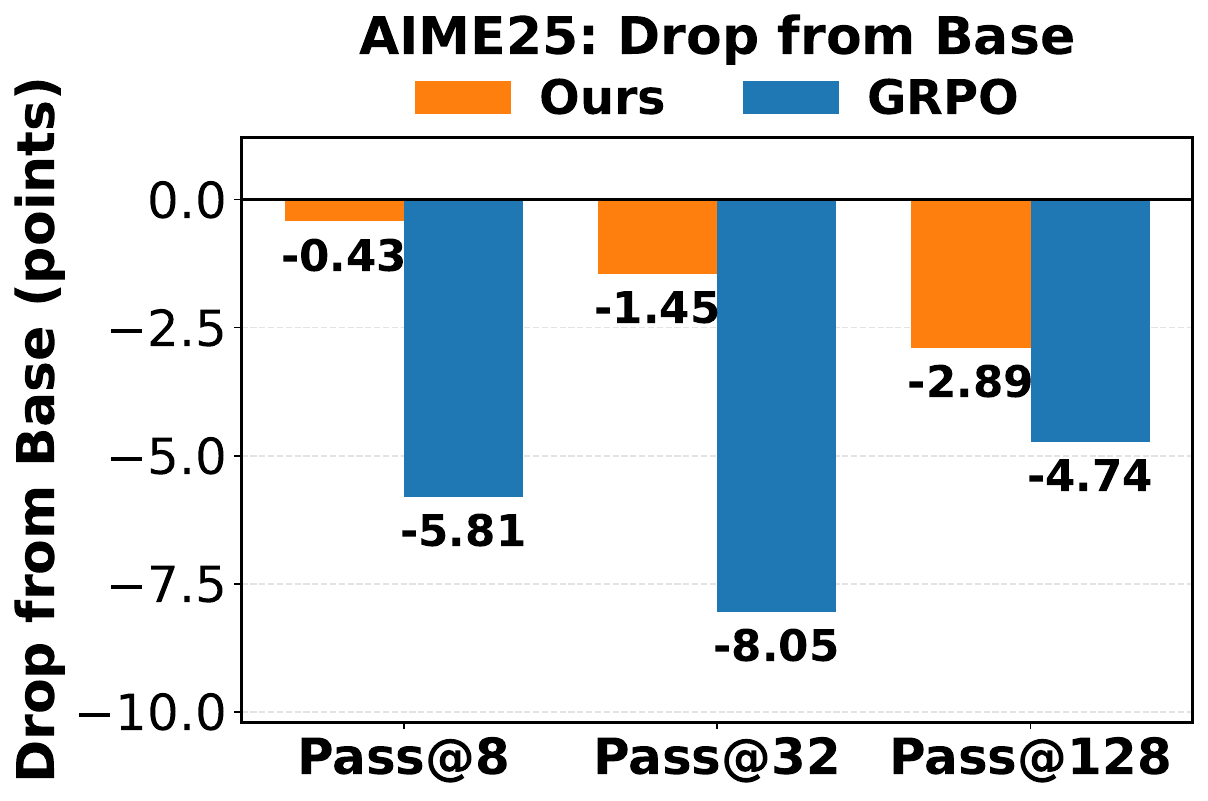}
    \captionsetup{width=\linewidth}
    \caption{
    Drop in Pass@$k$ relative to the base model on AIME25. Smaller drops indicate better preservation of multi-sampling performance after RL training.
    }
    \label{fig:drop_from_base_aime25}
\end{wrapfigure}

\subsection{Sampling-Based Evaluation}
Beyond greedy Pass@1, we further evaluate whether RL training preserves the model's exploration boundary under large sampling budgets. Prior work has reported that reasoning RL can improve single-sample accuracy while degrading Pass@$k$, suggesting a shrinkage of the model's effective capability frontier~\citep{yue2025does}. We therefore evaluate Pass@256 on AIME24 and AIME25 and report unbiased Pass@$k$ estimates from these samples. As shown in Figure~\ref{fig:drop_from_base_aime25}, our method yields a much smaller drop from the base model than GRPO on AIME25, indicating that it better preserves the target model's multi-sample exploration ability while still benefiting from RL training. The result on AIME24 is shown in Figure~\ref{fig:drop_24}.

\subsection{Ablation Study}
\begin{table}
\centering
\caption{
Greedy Pass@1 performance comparison across mathematical reasoning benchmarks.
All values are reported as percentages. The unbalanced ablation corresponds to the setting evaluated over 1,763 problems with one sample per problem.
}
\small
\begin{tabular}{lcccccc}
\toprule
\textbf{Model} &
\textbf{Overall} &
\textbf{MATH500} &
\textbf{Olympiad} &
\textbf{Omni-HARD} &
\textbf{AIME24} &
\textbf{AIME25} \\
\midrule
Qwen3-4B-Instruct-2507
& 58.3 & 90.8 & 64.4 & 21.2 & 50.0 & 40.0 \\

Unbalanced Ablation
& 62.6 & 94.0 & 69.7 & 25.7 & 46.7 & 46.7 \\

A-only Branch Ablation
& 61.9 & 93.2 & 70.6 & 23.3 & 53.3 & 36.7 \\

W2SPO (ours)
& \textbf{64.2} & 93.2 & \textbf{71.8} & \textbf{28.4} & \textbf{56.7} & \textbf{46.7} \\
\bottomrule
\end{tabular}
\label{tab:greedy_ablation_results}
\end{table}
\begin{figure}
    \centering
    \includegraphics[width=1\linewidth]{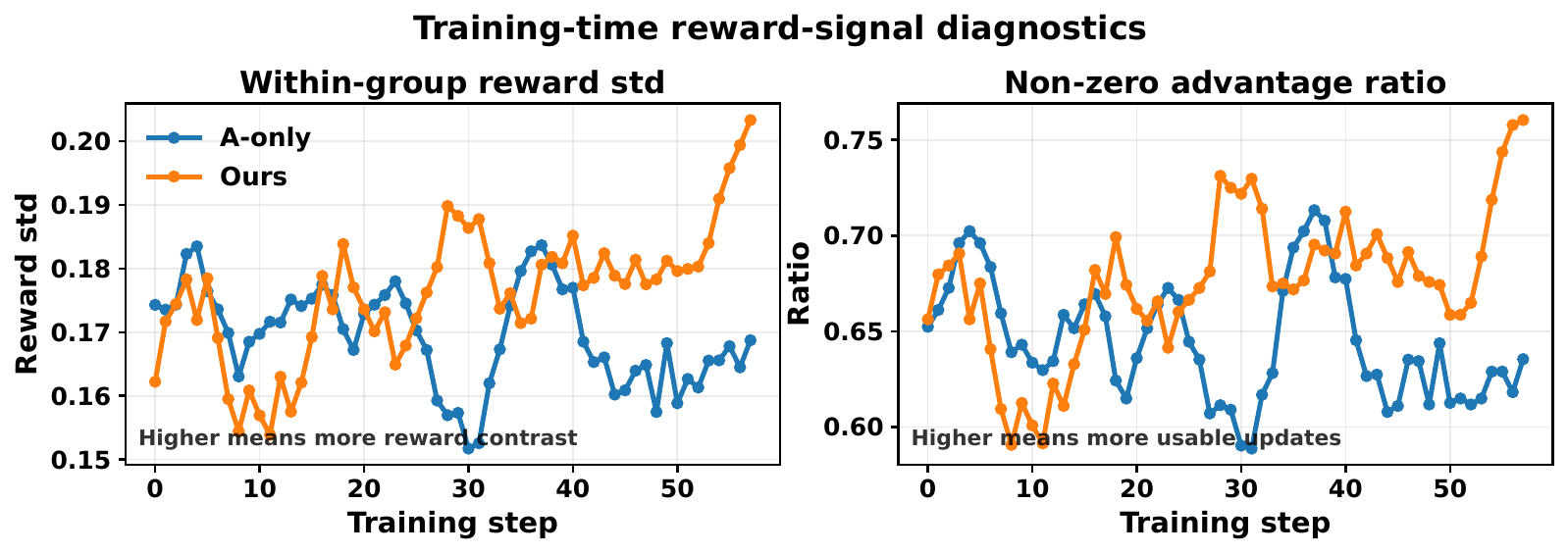}
    \caption{Training-time reward-signal diagnostics for the controlled A-only branch ablation.
}
    \label{fig:reward_diagnostics}
\end{figure}
Table~\ref{tab:greedy_ablation_results} reports the ablation results on greedy Pass@1 performance across six mathematical reasoning benchmarks.
% We compare our method with the base model, an A-only branch ablation that replaces all auxiliary-model branches with target-model branches, and an unbalanced branch-sampling variant.
% These comparisons test whether the gains come from reinforcement learning, cross-model branch diversity, and our branch-level importance-weighted advantage design.
% Our full method achieves the best overall performance.
Compared to the base model (Qwen3-4B-Instruct-2507), our method achieves a substantial improvement, boosting the overall accuracy from 58.3\% to 64.2\%. Notably, disabling cross-model branch diversity (the A-only Branch Ablation) leads to a performance drop to 61.9\% overall, validating the necessity of leveraging auxiliary models for branch diversity.  Furthermore, replacing our importance-weighted advantage design with the unbalanced variant (Unbalanced Ablation) results in a 1.6\% overall performance decline and a severe drop on AIME24 (from 56.7\% to 46.7\%). These results consistently demonstrate that both cross-model branch diversity and the importance-weighted advantage formulation are critical to the efficacy of the proposed reinforcement learning framework.

Figure~\ref{fig:reward_diagnostics} further compares our method with the A-only branch ablation under a controlled setting, where all hyperparameters, sampling configurations, and the random seed are kept identical except for replacing $B$-sampled branches with $A$-sampled branches. Our method yields a larger within-group reward standard deviation and a higher non-zero advantage ratio, indicating stronger reward contrast and more usable policy updates.

Robustness experiments further show that our method remains stable under both random-seed variation and auxiliary-model replacement.
We defer detailed results to Appendix~\ref{app:robustness_analysis}.

\section{Analysis: Local Credit Assignment and Importance-Aware Branch Weighting}
\label{sec:analysis}

We provide a local analysis to explain why short auxiliary branches can yield useful
training signals. This analysis is not a convergence guarantee for the full algorithm;
it considers a fixed-prefix, stop-gradient approximation that matches our implementation,
where rewards are obtained from completed rollouts but policy gradients are applied only
to the inserted branch tokens.

Let $x$ be a problem, $c$ a fixed actor-generated prefix, $z$ a short branch, and $y$
the actor continuation after the branch. We define the stop-gradient downstream value
\begin{equation}
    Q_{\bar{\theta}}(c,z)
    =
    \mathbb{E}_{y \sim \pi_{\bar{\theta}}(\cdot \mid x,c,z)}
    \left[ R(x,c,z,y) \right],
\end{equation}
and the local branch objective
\begin{equation}
    J_{\mathrm{br}}(\theta;c,\bar{\theta})
    =
    \mathbb{E}_{z \sim \pi_{\theta}(\cdot \mid x,c)}
    \left[ Q_{\bar{\theta}}(c,z) \right].
\end{equation}
Under this approximation,
\begin{equation}
    \nabla_{\theta} J_{\mathrm{br}}
    =
    \mathbb{E}_{z \sim \pi_{\theta}}
    \left[
    Q_{\bar{\theta}}(c,z)
    \nabla_{\theta}\log \pi_{\theta}(z \mid x,c)
    \right],
\end{equation}
where $\log \pi_{\theta}(z \mid x,c)$ decomposes over the $L_b$ branch tokens.
Thus, branch-token masking can be interpreted as optimizing a local decision whose
value is estimated by downstream verifier rewards, rather than treating the full
continuation as supervised data.

Auxiliary branches are off-policy proposals, so we do not use an exact unbiased
importance-sampling estimator. Instead, we use a tempered and clipped coefficient
$c_i$ to bound their influence in the branch-level advantage estimate. This allows
successful auxiliary proposals to affect the actor while preventing highly mismatched
branches from dominating the update. This analysis motivates three diagnostics:
higher within-group reward contrast, fewer all-wrong groups, and a reasonable effective
sample size under the balancing weights. Detailed assumptions and derivations are
provided in Appendix~\ref{app:local_analysis}.

\section{Conclusion and Limitations}

We introduced W2SPO, a weak-to-strong off-policy RL method for mathematical reasoning. 
Rather than distilling full auxiliary solutions, W2SPO uses a weaker frozen model to propose short local branches, lets the target model complete the reasoning trajectory, and applies verifier-based policy updates only to the inserted branch tokens. 
Experiments show that W2SPO improves over matched-sampling GRPO and target-only branching under the same evaluation protocol, achieving stronger greedy performance while better preserving multi-sampling capability. 
These results suggest that weak auxiliary branches can provide useful local exploration signals for strengthening the target reasoning policy.
Our current study is limited by computational resource constraints: the experiments are currently restricted to relatively small-scale language models and mathematical reasoning tasks.

\bibliographystyle{plainnat}
\bibliography{reference}

\appendix

\section{More Experimental Details and Results}
\subsection{Example Rollout}
\label{subsec:example_rollout}

We present one rollout group from W2SPO to illustrate how branch-level rewards are assigned.

The system prompt is :
\begin{quote}
\small
\texttt{Please reason step by step, and put your final answer within \textbackslash{}boxed\{\}.}
\end{quote}

The problem is:

\begin{quote}
\small
A $3$ by $2$ rectangle is split into four congruent right-angled triangles.
Those four triangles are rearranged to form a rhombus.
What is the ratio of the perimeter of the rectangle to the perimeter of the rhombus?
\end{quote}

The ground-truth answer is $1$. The target model $A$ first generates the following
prefix, after which short candidate branches are sampled from either the target model
itself or the auxiliary model $B$.

\begin{modelbox}{prefixblue}
\textbf{Target prefix generated by $A$:}

We are given:

- A 3 by 2 rectangle is split into four congruent right-angled triangles.

- These triangles are rearranged to form a rhombus.

- We are to find the ratio
\end{modelbox}

For this rollout group, we sample $n_A=2$ target branches and $n_B=6$ auxiliary
branches. Each candidate branch is inserted after the same prefix, and the target
model $A$ then completes the rest of the solution. Let $r_i$ be the final verifier
reward of branch $i$. We compute the importance-aware branch weight and advantage.

In this group, the rewards are
\[
[0,0,1,1,1,1,1,0],
\]
with weighted mean $\mu=0.625150$ and weighted standard deviation
$\sigma=0.484084$.

Below, we show one failed target branch. The
model-generated segments are highlighted in colored boxes.

\paragraph{Target branch: failed self-branch from $A$.}
For Branch 0, the inserted branch is sampled from the target model itself. The final
prediction is $\frac{\sqrt{10}}{4}$, so the verifier reward is $0$.

\begin{modelbox}{prefixblue}
\textbf{Prefix generated by $A$:}

We are given:

- A 3 by 2 rectangle is split into four congruent right-angled triangles.

- These triangles are rearranged to form a rhombus.

- We are to find the ratio
\end{modelbox}

\begin{modelbox}{abranchred}
\textbf{Inserted branch from $A$:}

of the perimeter of the rectangle to the
\end{modelbox}

\begin{modelbox}{contgray}
\textbf{Continuation generated by $A$:}

perimeter of the rhombus.

We'll solve this step by step.

Step 1: Understand the rectangle. The rectangle has dimensions length $3$ and width $2$,
so its area is $3\times 2=6$. Since it is split into four congruent right-angled
triangles, each triangle has area $6/4=1.5$.

The model then reasons that one possible split is to cut the rectangle into two
$3\times 1$ rectangles and cut each rectangle along a diagonal, giving right triangles
with legs $3$ and $1$ and hypotenuse $\sqrt{10}$. It then incorrectly treats the
rhombus perimeter as being determined by this hypotenuse-based construction and
outputs
\[
\boxed{\frac{\sqrt{10}}{4}}.
\]
\end{modelbox}

The branch-level scoring terms are:
\[
\log p_{\mathrm{old}}=-0.000738,\quad
\log q=-0.000738,\quad
c_i=1.000000,\quad
b_i=0.714384,\quad
A_i=-1.475743.
\]
Thus, this target self-branch receives a negative branch advantage.

\subsection{Sensitivity Analysis}
\label{sec:sensitivity_analysis}
\begin{figure}[h]
    \centering
    \includegraphics[width=1\linewidth]{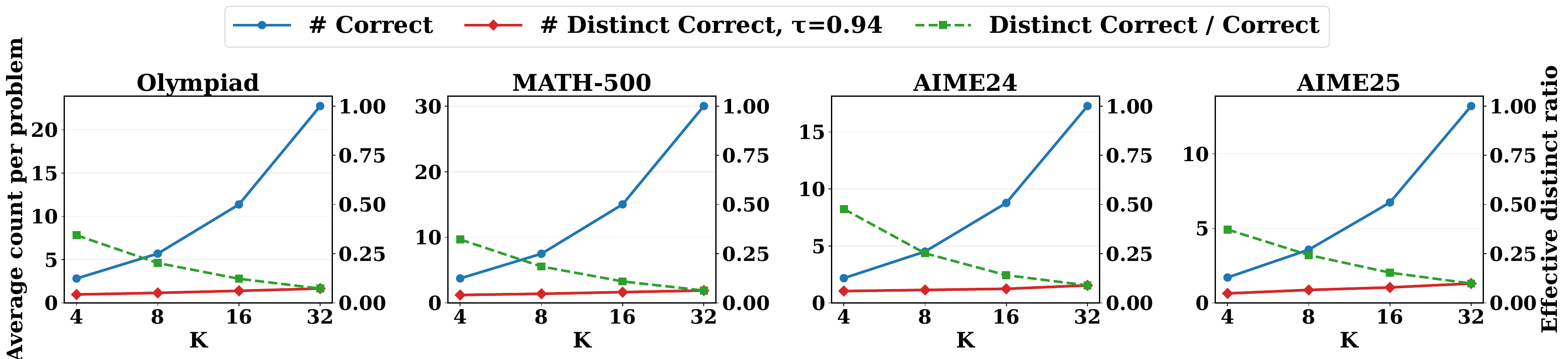}
    \caption{
    Semantic diversity of correct trajectories under similarity threshold $0.94$.
    }
    \label{fig:semantic_diversity_thr094}
\end{figure}

\begin{figure}
    \centering
    \includegraphics[width=1\linewidth]{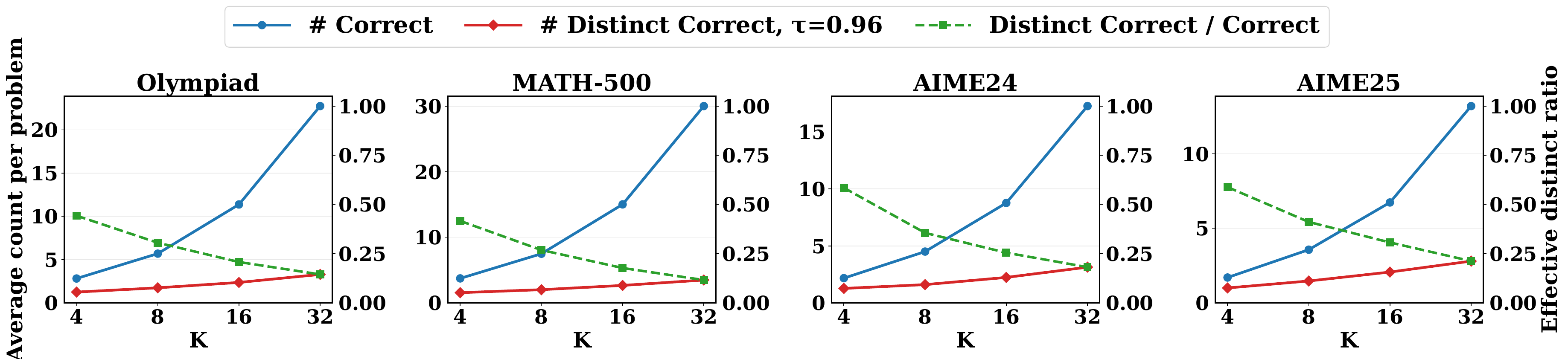}
    \caption{
    Semantic diversity of correct trajectories under similarity threshold $0.96$.
    }
    \label{fig:semantic_diversity_thr096}
\end{figure}
We further conduct a sensitivity analysis on the semantic similarity threshold used to identify distinct correct trajectories. 
Figures~\ref{fig:semantic_diversity_thr094} and~\ref{fig:semantic_diversity_thr096} report results under thresholds of $0.94$ and $0.96$, respectively. 
The overall trend remains consistent across both settings: auxiliary-branch trajectories produce more semantically distinct correct solutions than direct self-rollouts. 
This indicates that our motivation experiment is robust to reasonable choices of the semantic diversity threshold.
\subsection{Experimental Details}
\label{app:exp_details}
\textbf{Training Setup.} Training is implemented in an edited OpenRLHF pipeline with
Ray, vLLM, DeepSpeed ZeRO, bf16 precision, FlashAttention, and gradient
checkpointing. At each iteration, prompts are sampled from the training set,
converted with the tokenizer chat template, expanded into branch-level samples,
scored by a rule-based verifier, and used to update only the actor model.

\paragraph{Training data processing.}
We construct the training set from the mathematical training split of Skywork-OR1-RL-Data. 
To avoid benchmark contamination, we first remove problems that are identical or highly similar to examples in the evaluation benchmarks. 
From the remaining pool, we select problems whose difficulty scores under DeepSeek-R1-Distill-Qwen-32B fall between 12 and 15. 
We then evaluate Qwen3-4B-Instruct-2507 with eight sampled responses per problem and retain 2,956 problems with partial success, i.e., problems whose Pass@8 count is greater than 0 and smaller than 8. 
To further include challenging examples that the actor cannot solve under its current sampling distribution, we additionally select 800 problems with the shortest labels from the subset with Pass@8 count equal to 0. 
This yields a final training set of 3,756 mathematical reasoning problems.

\paragraph{Core training hyperparameters.}
For W2SPO training, we use Qwen3-4B-Instruct-2507 as the trainable actor and Gemma-3-4B-IT as a frozen auxiliary branch proposer. For each prompt, we construct a branch group with eight candidate branches: two sampled from the actor and six sampled from the auxiliary model. The intervention prefix length is set to $L_p=50$, and each inserted branch contains $L_b=8$ actor tokens. Each branch is evaluated with two actor continuations, with a maximum generation length of 8192 tokens. We train for two epochs using a learning rate of $1\times10^{-6}$, PPO clipping threshold $\epsilon=0.2$, and the importance-aware branch advantage estimator with $\alpha=0.02$, $c_{\max}=2.0$, and advantage clipping at 3.0. Rollouts are sampled with temperature 1.0 and top-$p$ 1.0. Training uses a rollout batch size of 64 and a training batch size of 96, with bf16 precision, FlashAttention, gradient checkpointing, and DeepSpeed ZeRO stage 2. In total, the setup uses two GPUs for actor training and six single-GPU vLLM engines for rollout generation.

\paragraph{Efficiency analysis.}
W2SPO is more efficient than matched-sampling GRPO because the policy-optimization
computation is localized to the short branch window. In standard GRPO, the actor
must compute log-probabilities and gradients over complete sampled trajectories,
whose lengths can be thousands of tokens in mathematical reasoning tasks. In
contrast, W2SPO only needs to run the actor forward over the prefix and the inserted
branch when computing branch-token log-probabilities, and the policy loss is applied
only to the $L_b=8$ branch tokens. Although complete continuations are still generated
to obtain verifier rewards, they are treated as downstream rollouts rather than tokens
to be optimized. As a result, both the forward computation for policy evaluation and
the backward computation for gradient updates are substantially reduced.

This localized update also reduces activation memory, since gradients are only
required for the short branch segment instead of the full reasoning trajectory. The
saved memory can be allocated to rollout generation, allowing larger effective memory
budgets for the vLLM engines. Consequently, under the same overall sampling budget,
W2SPO not only reduces the cost of policy optimization, but also slightly improves
rollout throughput. These factors together explain the large reduction in training
time observed in Table~\ref{tab:main_results_and_runtime}, especially in the forward and training
stages.
\subsection{Robustness Analysis}
\label{app:robustness_analysis}

To examine the robustness of our method, we evaluate two additional variants: one trained with a different random seed, and another replacing the default auxiliary weak model with \texttt{Mistral-7B-Instruct-v0.3}~\citep{jiang2023mistral7b}. 
As shown in Table~\ref{tab:robustness_analysis}, our method achieves consistently strong performance across both settings. 
Changing the random seed leads to only a small variation in overall performance, while replacing the auxiliary model still preserves most of the gains. 
These results suggest that the effectiveness of our method does not rely on a specific random seed or a particular auxiliary model, demonstrating its robustness.

\begin{table}[h]
\centering
\caption{
Robustness evaluation under different random-seed and auxiliary-model settings.
All results are reported as greedy Pass@1 percentages.
The different B-model variant replaces the default auxiliary model with \texttt{Mistral-7B-Instruct-v0.3}.
}
\small
\begin{tabular}{lcccccc}
\toprule
\textbf{Setting} &
\textbf{Overall} &
\textbf{MATH500} &
\textbf{Olympiad} &
\textbf{Omni-HARD} &
\textbf{AIME24} &
\textbf{AIME25} \\
\midrule
W2SPO (seed=42)
& \textbf{64.2} & 93.2 & \textbf{71.8} & \textbf{28.4} & \textbf{56.7} & 46.7 \\

W2SPO (seed=2026)
& 63.9 & \textbf{94.2} & 70.9 & 27.8 & 50.0 & \textbf{50.0} \\

Different B Model
& 63.4 & 93.8 & 70.8 & 27.4 & 50.0 & 40.0 \\
\bottomrule
\end{tabular}
\label{tab:robustness_analysis}
\end{table}

\begin{wrapfigure}[12]{r}{0.4\linewidth}
    \centering
    \vspace{-2.5em}
    \includegraphics[width=\linewidth]{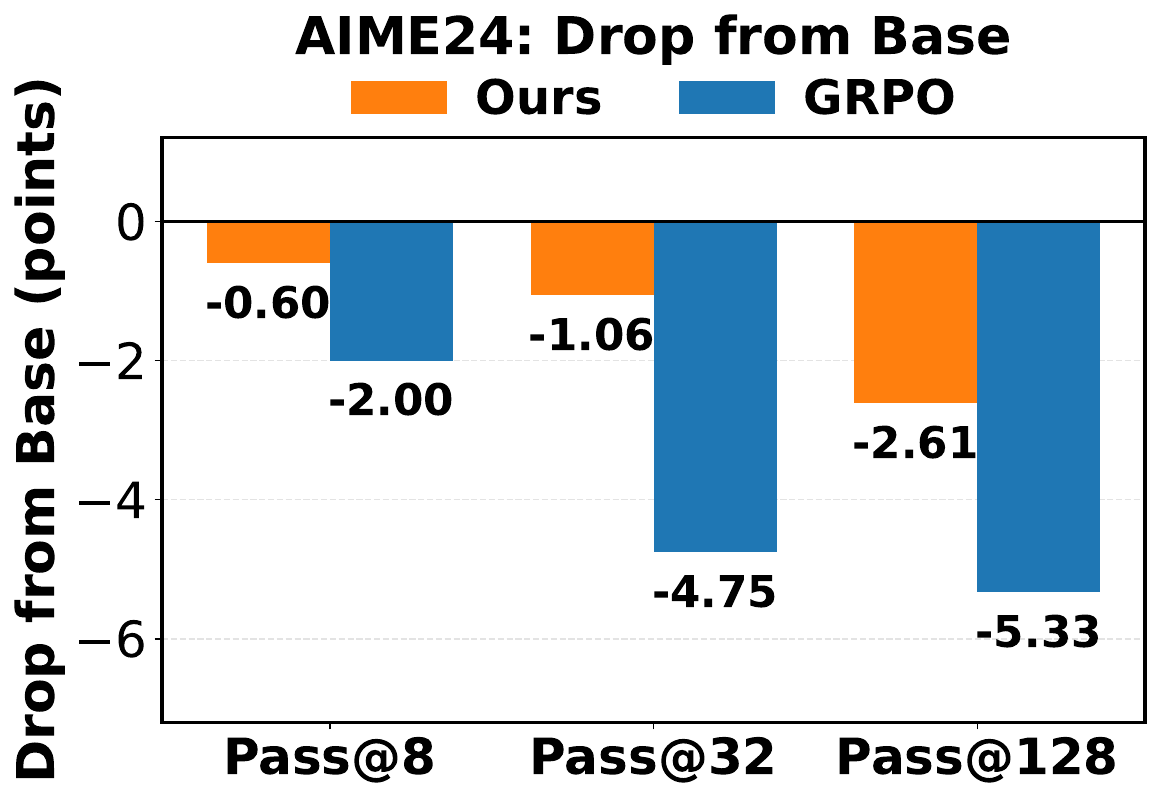}
    \caption{Drop in Pass@$k$ relative to the base model on AIME24}
    \label{fig:drop_24}
\end{wrapfigure}
\section{Additional Analysis of Branch-Level Updates}
\label{app:local_analysis}

This appendix gives a more explicit derivation of the local analysis in
Section~\ref{sec:analysis}. The purpose is to justify the interpretation of our update
as a branch-level credit-assignment mechanism. We do not claim that the resulting
algorithm is an unbiased estimator of the full off-policy policy gradient, nor do we
establish a convergence guarantee for the complete training procedure.

\paragraph{Setup.}
Fix a problem $x$ and a prefix $c$ generated by the actor. Let
$z=(z_1,\ldots,z_{L_b})$ be a short branch segment, and let $y$ be the continuation
generated by the actor after the branch. For the local analysis, the continuation policy
is treated as fixed. We denote this stop-gradient continuation policy by
$\pi_{\bar{\theta}}$, and define
\begin{equation}
    Q_{\bar{\theta}}(c,z)
    =
    \mathbb{E}_{y \sim \pi_{\bar{\theta}}(\cdot \mid x,c,z)}
    \left[
    R(x,c,z,y)
    \right].
\end{equation}
The local branch objective is
\begin{equation}
    J_{\mathrm{br}}(\theta;c,\bar{\theta})
    =
    \mathbb{E}_{z \sim \pi_{\theta}(\cdot \mid x,c)}
    \left[
    Q_{\bar{\theta}}(c,z)
    \right].
\end{equation}
This objective isolates the effect of changing the branch distribution while holding
the prefix and continuation-value estimator fixed.

\paragraph{Claim 1: Branch-token masking gives a local policy-gradient estimator under a fixed-prefix approximation.}
Assume that the support of $\pi_{\theta}(z \mid x,c)$ does not change with $\theta$
and that $Q_{\bar{\theta}}(c,z)$ is treated as independent of $\theta$ during the branch
update. Then
\begin{equation}
    \nabla_{\theta} J_{\mathrm{br}}(\theta;c,\bar{\theta})
    =
    \mathbb{E}_{z \sim \pi_{\theta}(\cdot \mid x,c)}
    \left[
    Q_{\bar{\theta}}(c,z)
    \nabla_{\theta}
    \log \pi_{\theta}(z \mid x,c)
    \right].
\end{equation}

\paragraph{Proof.}
By definition,
\begin{equation}
    J_{\mathrm{br}}(\theta;c,\bar{\theta})
    =
    \sum_z
    \pi_{\theta}(z \mid x,c) Q_{\bar{\theta}}(c,z).
\end{equation}
Taking the gradient and using
$\nabla_{\theta}\pi_{\theta}(z)=
\pi_{\theta}(z)\nabla_{\theta}\log\pi_{\theta}(z)$ gives
\begin{align}
    \nabla_{\theta} J_{\mathrm{br}}
    &=
    \sum_z
    \nabla_{\theta}\pi_{\theta}(z \mid x,c)
    Q_{\bar{\theta}}(c,z) \\
    &=
    \sum_z
    \pi_{\theta}(z \mid x,c)
    Q_{\bar{\theta}}(c,z)
    \nabla_{\theta}
    \log \pi_{\theta}(z \mid x,c) \\
    &=
    \mathbb{E}_{z \sim \pi_{\theta}}
    \left[
    Q_{\bar{\theta}}(c,z)
    \nabla_{\theta}
    \log \pi_{\theta}(z \mid x,c)
    \right].
\end{align}
Since $z$ is a token sequence,
\begin{equation}
    \log \pi_{\theta}(z \mid x,c)
    =
    \sum_{t=1}^{L_b}
    \log \pi_{\theta}(z_t \mid x,c,z_{<t}).
\end{equation}
Therefore, applying the policy-gradient loss only to the branch tokens corresponds to
optimizing this local branch objective. This is a semi-gradient view: it ignores the
dependence of future continuation probabilities on $\theta$, which is consistent with
the implementation choice of masking gradients to the branch window.

\paragraph{Baseline subtraction.}
For any baseline $b(c)$ that does not depend on the sampled branch $z$,
\begin{equation}
    \mathbb{E}_{z \sim \pi_{\theta}}
    \left[
    b(c)
    \nabla_{\theta}\log \pi_{\theta}(z \mid x,c)
    \right]
    =
    b(c)\nabla_{\theta}\sum_z \pi_{\theta}(z \mid x,c)
    =
    0.
\end{equation}
Thus, a centered branch value $Q_{\bar{\theta}}(c,z)-b(c)$ preserves the expected local
gradient while reducing variance. Our implementation uses a finite-sample, weighted
leave-one-out baseline and normalization. This should be viewed as a practical
variance-reduction and balancing heuristic, not as an exact unbiased estimator of the
full policy gradient.

\paragraph{Claim 2: Auxiliary branches can reduce all-wrong groups when they improve local success probability.}
Assume binary rewards $R \in \{0,1\}$. For a branch $z_i$, define its conditional
success probability as
\begin{equation}
    p_i
    =
    \Pr_{y \sim \pi_{\bar{\theta}}(\cdot \mid x,c,z_i)}
    \left[
    R(x,c,z_i,y)=1
    \right].
\end{equation}
If we draw $K$ independent continuations from the actor after this branch, then the
probability that branch $z_i$ obtains at least one successful continuation is
\begin{equation}
    s_i
    =
    1-(1-p_i)^K.
\end{equation}
This quantity is monotone in both $p_i$ and $K$.

If success events are further approximated as independent across branches, then for a
group of $G$ branches,
\begin{equation}
    P_{\mathrm{all\text{-}wrong}}
    =
    \prod_{i=1}^{G}(1-s_i).
\end{equation}
Therefore, under this independence approximation, replacing actor-only branches with
auxiliary branches that have larger conditional success probabilities reduces the
all-wrong probability multiplicatively. This statement is conditional: it does not
assume that every auxiliary branch is useful, nor that the auxiliary model is a stronger
standalone solver. It only requires that some auxiliary branches induce larger downstream
success probabilities under the actor continuation policy.

Without the independence approximation, we can still state a weaker bound. Since the
event that all branches fail is a subset of the event that any particular branch $i$
fails,
\begin{equation}
    P_{\mathrm{all\text{-}wrong}}
    \le
    1-s_i
    \quad
    \text{for every } i.
\end{equation}
Hence,
\begin{equation}
    P_{\mathrm{all\text{-}wrong}}
    \le
    1-\max_i s_i.
\end{equation}
This bound shows that a single branch with high downstream success probability can
lower an upper bound on the all-wrong probability, although the multiplicative formula
requires the stronger independence approximation.

\paragraph{Off-policy branch proposals.}
In our method, branches are sampled from both the actor and an auxiliary proposal
distribution. If all samples came from a known mixture proposal
\begin{equation}
    m(z \mid x,c)
    =
    \lambda \pi_{\theta_{\mathrm{old}}}(z \mid x,c)
    +
    (1-\lambda)q_B(z \mid x,c),
\end{equation}
then an exact importance-sampling correction for estimating expectations under a
target branch policy would involve a ratio of the form
\begin{equation}
    w^*(z)
    =
    \frac{\pi_{\theta_{\mathrm{old}}}(z \mid x,c)}
    {m(z \mid x,c)}
    \quad
    \text{or}
    \quad
    \frac{\pi_{\theta}(z \mid x,c)}
    {m(z \mid x,c)},
\end{equation}
depending on the target distribution. Such ratios can have high variance and are
difficult to compute exactly when auxiliary text is decoded and re-tokenized into the
actor vocabulary.

Instead, we use a source-specific, tempered, and clipped coefficient
\begin{equation}
    c_i
    =
    \min \left\{
    c_{\max},
    \exp \left(
    \alpha \cdot
    \mathrm{clip}
    \left(
    \ell_i - \eta_i,
    -M, M
    \right)
    \right)
    \right\},
\end{equation}
where
\begin{equation}
    \ell_i
    =
    \log \pi_{\theta_{\mathrm{old}}}(z_i \mid x,c)
\end{equation}
is the old actor log-probability of the branch under the actor tokenizer, and
\begin{equation}
    \eta_i
    =
    \log q_i(z_i \mid x,c)
\end{equation}
is the source proposal log-probability used for that branch. For actor-sampled branches,
$q_i$ is the actor proposal distribution. For auxiliary-sampled branches, $q_i$ denotes
the auxiliary proposal score after mapping the generated text to the actor-side branch
representation.

Because of tempering, clipping, source-specific proposals, and tokenizer conversion,
$c_i$ is not an unbiased importance-sampling ratio. Its purpose is instead to provide a
bounded bias--variance tradeoff. The coefficient satisfies
\begin{equation}
    0 < c_i \le c_{\max},
\end{equation}
so a branch cannot receive an arbitrarily large scalar weight from the balancing term.
Together with advantage clipping and PPO ratio clipping, this prevents highly off-policy
auxiliary branches from dominating the branch-token update. This does not bound the full
parameter-space gradient norm, but it bounds the scalar weighting used inside the
surrogate objective.

\paragraph{Effective sample size.}
To monitor whether the balancing weights concentrate on a small number of branches, we
use the effective sample size
\begin{equation}
    \mathrm{ESS}
    =
    \frac{\left(\sum_i c_i\right)^2}
    {\sum_i c_i^2}.
\end{equation}
For a group of $G$ branches, one may also report the normalized effective sample size
\begin{equation}
    \mathrm{nESS}
    =
    \frac{1}{G}
    \frac{\left(\sum_i c_i\right)^2}
    {\sum_i c_i^2}.
\end{equation}
A very small nESS indicates that the update is dominated by a few weighted branches,
whereas a moderate nESS suggests that auxiliary proposals contribute without collapsing
the effective training group.

\paragraph{Empirical diagnostics.}
The analysis motivates, but does not prove, the following diagnostics. If auxiliary
branches provide useful local exploration, we expect to observe:
\begin{enumerate}
    \item higher within-group reward contrast compared with actor-only branch sampling;
    \item lower all-wrong ratio and higher mixed-correct ratio;
    \item moderate ESS or nESS, together with a controlled clipping fraction.
\end{enumerate}
These diagnostics test whether auxiliary branches improve local exploration while
remaining sufficiently aligned with the actor distribution.
\input{checklist.tex}
\end{document}

%% file: checklist.tex
\section*{NeurIPS Paper Checklist}

\begin{enumerate}

\item {\bf Claims}
    \item[] Question: Do the main claims made in the abstract and introduction accurately reflect the paper's contributions and scope?
    \item[] Answer: \answerYes{} % Replace by \answerYes{}, \answerNo{}, or \answerNA{}.
    \item[] Justification: The abstract and introduction state the main claims of W2SPO, including the support-limited failure mode of self-rollout RL, the auxiliary-branch method, and the empirical improvements. The scope is limited to mathematical reasoning tasks and is further discussed in the limitations section.

\item {\bf Limitations}
    \item[] Question: Does the paper discuss the limitations of the work performed by the authors?
    \item[] Answer: \answerYes{} % Replace by \answerYes{}, \answerNo{}, or \answerNA{}.
    \item[] Justification: The paper discusses limitations in the conclusion, including the focus on relatively small-scale language models and mathematical reasoning tasks due to computational resource constraints.

\item {\bf Theory assumptions and proofs}
    \item[] Question: For each theoretical result, does the paper provide the full set of assumptions and a complete (and correct) proof?
    \item[] Answer: \answerYes{} % Replace by \answerYes{}, \answerNo{}, or \answerNA{}.
    \item[] Justification: The analysis section explicitly states that the derivation is a local fixed-prefix, stop-gradient approximation rather than a convergence guarantee. Additional assumptions and derivations are provided in the appendix.

    \item {\bf Experimental result reproducibility}
    \item[] Question: Does the paper fully disclose all the information needed to reproduce the main experimental results of the paper to the extent that it affects the main claims and/or conclusions of the paper (regardless of whether the code and data are provided or not)?
    \item[] Answer: \answerYes{} % Replace by \answerYes{}, \answerNo{}, or \answerNA{}.
    \item[] Justification: The paper describes the model initialization, training data processing, branch configuration, reward computation, evaluation benchmarks, and verification protocol. Additional hyperparameters and implementation details are provided in the appendix and anonymized code release.


\item {\bf Open access to data and code}
    \item[] Question: Does the paper provide open access to the data and code, with sufficient instructions to faithfully reproduce the main experimental results, as described in supplemental material?
    \item[] Answer: \answerYes{} % Replace by \answerYes{}, \answerNo{}, or \answerNA{}.
    \item[] Justification: The paper provides an anonymized repository containing code and data for reproducing the main experiments. The repository is anonymized for the submission stage.


\item {\bf Experimental setting/details}
    \item[] Question: Does the paper specify all the training and test details (e.g., data splits, hyperparameters, how they were chosen, type of optimizer) necessary to understand the results?
    \item[] Answer: \answerYes{} % Replace by \answerYes{}, \answerNo{}, or \answerNA{}.
    \item[] Justification: The paper specifies the training setup, data filtering procedure, model choices, branch lengths, sampling configuration, optimization hyperparameters, and evaluation protocol.


\item {\bf Experiment statistical significance}
    \item[] Question: Does the paper report error bars suitably and correctly defined or other appropriate information about the statistical significance of the experiments?
    \item[] Answer: \answerYes{} % Replace by \answerYes{}, \answerNo{}, or \answerNA{}.
    \item[] Justification: For the runtime and speedup experiments, we report mean values with standard divisions computed over repeated training steps, which supports the efficiency claims. For benchmark accuracy, we report point estimates under a fixed evaluation protocol and include robustness checks with a different random seed and auxiliary model.

\item {\bf Experiments compute resources}
    \item[] Question: For each experiment, does the paper provide sufficient information on the computer resources (type of compute workers, memory, time of execution) needed to reproduce the experiments?
    \item[] Answer:\answerYes{} % Replace by \answerYes{}, \answerNo{}, or \answerNA{}.
    \item[] Justification: The paper reports the GPU setup and runtime measurements. The appendix further specifies that training uses two GPUs for actor training and six single-GPU vLLM engines for rollout generation.

    
\item {\bf Code of ethics}
    \item[] Question: Does the research conducted in the paper conform, in every respect, with the NeurIPS Code of Ethics \url{https://neurips.cc/public/EthicsGuidelines}?
    \item[] Answer: \answerYes{} % Replace by \answerYes{}, \answerNo{}, or \answerNA{}.
    \item[] Justification: The research uses publicly available models, datasets, and mathematical reasoning benchmarks, and does not involve human subjects, private data, or harmful data collection. The submission is anonymized in accordance with NeurIPS policy.


\item {\bf Broader impacts}
    \item[] Question: Does the paper discuss both potential positive societal impacts and negative societal impacts of the work performed?
    \item[] Answer: \answerYes{} % Replace by \answerYes{}, \answerNo{}, or \answerNA{}.
    \item[] Justification: The work may improve the reasoning ability and efficiency of language models, which can benefit mathematical problem solving and scientific assistance. As with general improvements to LLM reasoning, it may also increase the capability of systems that could be misused, so responsible deployment and evaluation remain important.

    
\item {\bf Safeguards}
    \item[] Question: Does the paper describe safeguards that have been put in place for responsible release of data or models that have a high risk for misuse (e.g., pre-trained language models, image generators, or scraped datasets)?
    \item[] Answer: \answerNA{} % Replace by \answerYes{}, \answerNo{}, or \answerNA{}.
    \item[] Justification:  The paper does not release a high-risk pretrained model, scraped dataset, or system intended for direct deployment. The released assets are code and mathematical reasoning data for research reproduction.


\item {\bf Licenses for existing assets}
    \item[] Question: Are the creators or original owners of assets (e.g., code, data, models), used in the paper, properly credited and are the license and terms of use explicitly mentioned and properly respected?
    \item[] Answer: \answerYes{} % Replace by \answerYes{}, \answerNo{}, or \answerNA{}.
    \item[] Justification: The paper cites the existing models, datasets, benchmarks, and software tools used in the experiments. The released repository documents the corresponding sources and usage requirements.


\item {\bf New assets}
    \item[] Question: Are new assets introduced in the paper well documented and is the documentation provided alongside the assets?
    \item[] Answer: \answerYes{}% Replace by \answerYes{}, \answerNo{}, or \answerNA{}.
    \item[] Justification: The paper releases anonymized code and processed data to support reproducibility. Documentation and usage instructions are provided in the anonymized repository.


\item {\bf Crowdsourcing and research with human subjects}
    \item[] Question: For crowdsourcing experiments and research with human subjects, does the paper include the full text of instructions given to participants and screenshots, if applicable, as well as details about compensation (if any)? 
    \item[] Answer: \answerNA{}% Replace by \answerYes{}, \answerNo{}, or \answerNA{}.
    \item[] Justification: The paper does not involve crowdsourcing experiments or research with human subjects.


\item {\bf Institutional review board (IRB) approvals or equivalent for research with human subjects}
    \item[] Question: Does the paper describe potential risks incurred by study participants, whether such risks were disclosed to the subjects, and whether Institutional Review Board (IRB) approvals (or an equivalent approval/review based on the requirements of your country or institution) were obtained?
    \item[] Answer: \answerNA{} % Replace by \answerYes{}, \answerNo{}, or \answerNA{}.
    \item[] Justification: The paper does not involve human subjects, so IRB approval or equivalent review is not applicable.


\item {\bf Declaration of LLM usage}
    \item[] Question: Does the paper describe the usage of LLMs if it is an important, original, or non-standard component of the core methods in this research? Note that if the LLM is used only for writing, editing, or formatting purposes and does \emph{not} impact the core methodology, scientific rigor, or originality of the research, declaration is not required.
    %this research? 
    \item[] Answer: \answerYes{} % Replace by \answerYes{}, \answerNo{}, or \answerNA{}.
    \item[] Justification: LLMs are central to the research method: the trainable actor and the frozen auxiliary model are both language models used for reasoning rollouts and auxiliary branch proposals. Their roles are described in the method and experimental setup.

\end{enumerate}